\pdfoutput=1

\documentclass[11pt]{article}

\usepackage[preprint]{acl}

\usepackage{times}
\usepackage{latexsym}

\usepackage[T1]{fontenc}

\usepackage[utf8]{inputenc}

\usepackage{microtype}

\usepackage{inconsolata}

\usepackage{graphicx}
\usepackage{listings}
\usepackage{booktabs}
\usepackage{multirow}

\usepackage{amsfonts,amsmath,mathtools}
\usepackage{xcolor}
\title{360-LLaMA-Factory: Plug \& Play Sequence Parallelism for Long Post-Training}

\author{\textbf{Haosheng Zou}$^{*3}$, \textbf{Xiaowei Lv}$^{*2}$, \textbf{Shousheng Jia}$^1$, \textbf{Lin Li}$^1$, \textbf{Xiaochun Gong}$^1$, \textbf{Xiangzheng Zhang}$^1$ \\
$^1$Qiyuan Tech \quad $^2$Renmin University \quad $^3$work done at Qiyuan Tech, now at ByteDance\\
\texttt{zouhaosheng@163.com} \\
*co-first authors}

\begin{document}
\maketitle
\begin{abstract}
Adding sequence parallelism into LLaMA-Factory, we open-sourced 360-LLaMA-Factory at \url{https://github.com/Qihoo360/360-LLaMA-Factory}.
360-LLaMA-Factory has received wide recognition and used in models such as Light-R1, TinyR1, Kaggle AIMO math models and also in large companies' training frameworks.
This technical report delves deeper into the different sequence parallel modes behind 360-LLaMA-Factory and discusses our implementation insights.

Large Language Models (LLMs) nowadays are expected to process inputs of unprecedented length, with current token limits extending to several million tokens.
To meet the long-sequence demands of LLMs, fine-tuning frameworks must also support post-training on extended sequences.
Based on the LLaMA-Factory framework, we implemented multiple sequence parallelism (DeepSpeed-Ulysses and Ring-Attention), provided feasible support for sequence parallelism of long sequences. 
Meanwhile, we extended DeepSpeed-Ulysses by adding dummy heads to handle cases where the number of attention heads is not divisible by the sequence parallel size.
At the same time, we conducted an in-depth analysis of the practical issues and potential errors of applying sequence parallelism to post-training.
Finally, we experimentally validated the correctness of our sequence parallelism implementation and demonstrated the efficiency of our Dummy-Head Ulysses. We also compared different sequence parallel strategies in terms of maximum sequence length and runtime efficiency. 
\end{abstract}

\section{Introduction}
Large language models (LLMs) are becoming increasingly important for long sequence performance, and the context length of today’s state-of-the-art models \cite{liu2024deepseek,yang2024qwen2} has reached millions of tokens.
Although computing speed has been improved, limited computing resources sometimes become a bottleneck for training. To achieve post-training of long sequences, sequence parallelism has become a necessity. Ring-Attention \cite{liu2023ring} and DeepSpeed-Ulysses \cite{jacobs2023deepspeed} are the two most common implementations of sequence parallelism. Although they have been applied to some existing frameworks \cite{2023xtuner,zhao2025swift,hu2024openrlhf}, some implementations have problems, and there is still no work that has fully explored the specific implementation details of sequence parallelism and the comparison of the details of different sequence parallelism.

Among the existing open source training frameworks, the LLaMA-Factory \cite{zheng2024llamafactory} framework supports many models and various training functions, but unfortunately, it does not support sequence parallelism. Based on the LLaMA-Factory framework, we implemented the sequence parallelism of Ring-Attention and Deepspeed-Ulysses, and only one line of extra code is needed to implement sequence parallel post-training. At the same time, we try to allow sequence parallelism to coexist with most of the original functions and optimizations to ensure that a series of functions such as LoRA \cite{hu2022lora} and neat-packing \cite{kundu2024enhancing} can be compatible normally.

In addition, we noticed that the DeepSpeed-Ulysses must satisfy the requirement that the number of attention heads is divisible by sequence parallel size. One possible approach is to combine DeepSpeed-Ulysses with other sequence parallelism methods, as demonstrated by USP \cite{fang2024usp} which integrates Ring-Attention, and LoongTrain \cite{gu2024loongtrain} which incorporates Double-Ring Attention. However, incorporating other sequence parallelism mechanisms may lose the efficiency advantages of DeepSpeed-Ulysses.
Another approach, as adopted by Xtuner \cite{2023xtuner}, avoids integrating with other sequence parallelism methods. Instead, it creates virtual attention heads by partitioning the hidden dimension. However, Xtuner's solution not only takes up more memory, but also reduces computational efficiency. We proposed a simple and effective supplementary solution, Dummy-Head Ulysses. By making up for the empty dummy heads, we can reduce the additional overhead and make the sequence parallel size of ulysses no longer limited by the number of heads. We have proved through experiments that this solution has significant improvements in memory optimization and efficiency optimization compared to Xtuner's implementation.


In addition, we conducted systematic analysis of critical challenges in sequence parallelism, including distributed communication problem and the impact of position IDs. Furthermore, we performed comprehensive comparsion between different sequence parallel implementations across key metrics, including throughput and maximum sequence length capacity, providing empirical foundations for practical deployment decisions.


Our contributions are summarized in the following three points:
\begin{itemize}
    \item Based on the LLaMA-Factory framework, we implemented the post-training sequence parallel functions (SFT and DPO) of Ring-Attention and DeepSpeed-Ulysses, while taking into account most of the original functions and providing support for post-training of long sequences. The correctness of our implementation is proved through experimental loss error analysis.
    \item We proposed a new method of adding empty dummy heads to make up for the shortcomings of ulysses, and verified its memory and efficiency improvements through experiments.
    \item We explored the specific issues of applying different sequence parallelism to actual post-training, and compared the maximum training length and efficiency of different sequence parallel methods.
    
\end{itemize}

\section{Related Work}

\subsection{Sequence Parallelsim}
Due to the need for long sequence training, sequence parallelism is becoming increasingly important. Deepspeed-Ulysses \cite{jacobs2023deepspeed} and Ring-Attention \cite{liu2023ring} are two important sequence parallel technologies. The former converts the sequence parallelism in attention calculation into head parallelism through all-to-all communication, while the latter's attention output is obtained by iterative calculation of local query chunk with all KV chunks. The communication of Ring-Attention is organized in a ring form, and each GPU sends and receives KV chunks at the same time. The biggest problem with Ulysses is that it must satisfy the requirement that the sequence parallel size is divisible by the head num, so for certain models, it may be difficult to support the 8 GPUs sequence parallel training.
USP \cite{fang2024usp} combines Ulysses and Ring-Attention and proposes a new Unified Sequence Parallelism Attention. In addition, LoongTrain \cite{gu2024loongtrain} also pays attention to the scalability and efficiency issues of existing methods, proposes a new 2D attention mechanism, combines head parallelism and sequence parallelism, and proposes Double-Ring attention to accelerate training. Although the technology of sequence parallelism has become mature, the details of its application to actual training frameworks still need to be explored.

\subsection{Fine-Tuning Frameworks}
Due to the increasing demand for LLM fine-tuning tasks, more and more LLM fine-tuning frameworks have emerged. Megatron-LM \cite{shoeybi2019megatron} serves as a research-oriented framework leveraging Megatron-Core for LLM training. OpenRLHF \cite{hu2024openrlhf} is a high-performance Reinforcement Learning from Human Feedback (RLHF)  \cite{ouyang2022training} framework built on Ray \cite{moritz2018ray}, DeepSpeed \cite{rasley2020deepspeed}, vLLM \cite{kwon2023efficient} and Hugging Face Transformers \cite{wolf2020transformers}. Ms-swift \cite{zhao2025swift} is an official framework for fine-tuning and deploying large language models and multi-modal large models, which supports the training, inference, evaluation, quantization, and deployment. XTuner \cite{2023xtuner} is an efficient, flexible and full-featured toolkit for fine-tuning large models, which supports continuous pre-training, instruction fine-tuning, and agent fine-tuning. Although the above frameworks all implement sequence parallel functions, these frameworks may suffer from usability challenges, complex code encapsulation, and even errors in sequence parallelism. LLaMA-Factory \cite{zheng2024llamafactory} is a unified framework that integrates a suite of efficient training methods. Although it is fully functional and easy to use, it does not yet implement sequence parallel capabilities. Therefore, it is necessary to provide support for sequence parallelism in LLaMA-Factory and conduct systematic research on the practical application of sequence parallelism.



\section{Developing Sequence Parallelism}
We first describe the specific implementation of sequence parallelism, focusing on the declaration of sequence parallel groups, data processing, and the post-training loss calculation processing. 

\subsection{Initialization}





The initialization of sequential parallelism includes grouping of corresponding GPUs and attention replacement. Given the sequence parallel size $sp$ ($sp>1$) and gpu nums $N$, each group contains all $N / sp$ GPUs, which subsequently support communication between GPUs. 

Following the initialization of communication groups, we employ a monkey patch to substitute the default attention function with either Ring-Attention or DeepSpeed-Ulysses. The Ring-Attention implementation is adapted from the \textit{zigzag\_ring\_flash\_attn\_func} provided by the \textit{ring-flash-attention} \footnote{\url{https://github.com/zhuzilin/ring-flash-attention}} library, whereas the DeepSpeed-Ulysses variant is modified form the \textit{UlyssesAttention} function in the \textit{yunchang} \footnote{\url{https://github.com/feifeibear/long-context-attention/}} library. It is worth noting that the operations we described above are all performed before loading the model.

\subsection{Data Processing}
Since sequence parallelism requires data to be split onto GPUs in a same sequence parallel group for parallel computing, we need to preprocess the data. First, we pad the input sequences to a length divisible by \textit{8 × sequence parallel size}. In practice, we further pad the sequences to the length closest to the \textit{cutoff\_len} (which is the maximum input length) parameter that satisfies this constraint. Subsequently, all fields in the input data are evenly partitioned into multiple segments according to the sequence parallel size. For DeepSpeed-Ulysses, a simple sequential split of the data is sufficient. But for Ring-Attention, in order to achieve load balancing of multi-GPUs computing, we need to use zigzag split \footnote{\url{https://github.com/zhuzilin/ring-flash-attention/issues/2}} \cite{fang2024usp}.
It is worth noting that in order to make DeepSpeed-Ulysses compatible with the neat-packing function, we need to keep the \textit{attention\_mask} without splitting it, but copy it to other GPUs in the sequence parallel group for subsequent processing.

\subsection{Correct Loss Calculation}
In Supervised Fine-Tuning, the model is fine-tuned via supervised learning, optimizing a loss function based on human-provided labels, as shown in Equation \ref{eq:sftloss}:
\begin{equation}
  \label{eq:sftloss}
  L_\theta=-\sum_i \log{p_\theta}(x_i | inst, x_{< i})
\end{equation}
where $\theta$ is the parameter of the model, $x_i$ is the $i_{th}$ token in the sequence, and $inst$ is the human instructions.

However, in the implementation of sequence parallelism, since only local output results are calculated on each GPU, the loss is only partial loss of part of sequence. Therefore, the final calculation loss should be performed by \textit{all\_reduce} operation to sum.

Directed Preference Optimization (DPO) is used to train the model to fit human preferences, whose specific loss function is shown in Equation \ref{eq:dpoloss}:

\begin{equation}
\begin{split}
  \label{eq:dpoloss}
  &\mathcal{L}_{\mathrm{DPO}}(\pi_\theta; \pi_{ref}) = 
  -\mathbb{E}_{(x, y_w, y_l)} 
  \\ 
  &\bigg[ \log{\sigma}\bigg(\beta \log{\frac{\pi_\theta(y_w|x)}{\pi_{ref}(y_w|x)}-\beta\frac{\pi_\theta(y_l|x)}{\pi_{ref}(y_l|x)}}\bigg) \bigg]
  \end{split}
\end{equation}

where $x$ is the prompt, $y_w$ and $y_l$ denotes the preferred and dispreferred completion, $\pi_\theta$ and $\pi_{ref}$ denotes the policy model and reference model respectively, the $\beta$ is a hyperparameter and the $\sigma$ denotes the sigmoid \cite{han1995influence} function.

Due to the impact of sequence parallelism, it is necessary to perform an \textit{all-reduce} operation across GPUs within the same sequence parallel group to obtain the final loss. However, unlike in SFT, the presence of the sigmoid function prevents direct \textit{all-redue} on loss. Instead, we first perform \textit{all-reduce} operations on the $\pi_\theta(y_w|x)$, $\pi_{ref}(y_w|x)$, $\pi_\theta(y_l|x)$, $\pi_{ref}(y_l|x)$ respectively, and then compute the DPO loss according to Equation \ref{eq:dpoloss}.

\subsection{SP on Vision-Language Models}

To enable sequence parallelism on multimodal models, we address the challenge that visual modalities (images and videos) only have embeddings without corresponding token IDs, unlike text token IDs that can be directly partitioned. Our approach adopts a unified framework where visual content is represented using placeholder image token IDs in the input sequence, with the number of placeholders predetermined by the multimodal processor based on image characteristics. 

Specifically, we construct an accompanying image position map of identical length to the input sequence, where each position contains either -1 for non-visual tokens or the corresponding visual token index (0, 1, 2, ...) for image placeholders. Both the input token sequence and position map undergo identical sequence-parallel partitioning, either in a zigzag splitting pattern (zigzag-ring) or naive splitting (ulysses). During the forward pass, we employ targeted monkey patching to utilize the position map for retrieving corresponding visual embeddings from ViT outputs, effectively replacing the placeholder tokens. 

This design maintains consistency with existing text-based sequence parallelism while seamlessly extending support to multimodal scenarios. Additionally, considering the significant variation in sequence lengths within multimodal batches, we implement an adaptive padding strategy that only pads sequences to the minimum required length, rather than using a fixed maximum length. This optimization reduces approximately 50\% of training time on our datasets by minimizing computational overhead from excessive padding.

\section{Dummy Head Ulysses}
There is a problem with DeepSpeed-Ulysses. Since it converts the sequence parallelism into the head parallelism in the attention through an all-to-all operation, it cannot handle the situation where the head nums in the attention cannot divide the sequence parallel size.

Suppose the input sequence before attention is $[bs, seq\_len / sp, hs, dim]$, where 
$bs$ is the batch size, $seq\_len$ is the input sequence length, $sp$ is the sequence parallel size, $hs$ is the head num, and $dim$ is hidden\_dim after multi-head dimensional transformation. After the all-to-all operation, it will be converted to $[bs, seq\_len, hs / sp, dim]$. When $hs$ is not divisible by sp, the above operation will fail. 

Xtuner \cite{2023xtuner} addresses this issue by transforming the hidden dimension to introduce additional virtual heads, making it divisible. This approach involves a dimensional transformation, converting the shape from $[bs, seq\_len / sp, hs, dim]$ to $[bs, seq\_len / sp, insp \times hs, dim / insp]$, thereby expanding the number of virtual heads. Here, $insp$ represents an additional internal communication sequence parallel group, which aggregates the hidden dimension through an internal $all\_gather$ operation. This transformation reshapes the sequence into $[bs, seq\_len, hs \times insp, dim]$, followed by additional attention operations. As a result, the hidden dimension is effectively recalculated $insp$ times during the actual computation. However, in practice, this approach incurs higher memory consumption and increased communication overhead. To solve this problem, we use a simple method to add a few empty heads to solve this problem.

\subsection{Dummy Head Implementation}
If the number of attention heads is not divisible by the sequence parallel size $sp$, we pad the head dimension by adding $sp - (hs \% sp)$ additional heads, so that the total number of heads becomes divisible by $sp$. To ensure correctness during both the forward and backward passes, we extend the input along the head dimension before the all-to-all operation, resulting in an input shape of $[bs, seq\_len / sp, hs_{new}, dim]$, where $hs_{new}$ includes the padded heads. The extra heads are then removed appropriately during the backward pass. The corresponding code is shown in Algorithm \ref{dummy_head}:

\begin{lstlisting}[caption={Dummy-Head-Ulysses}, label={dummy_head}]
def pad_heads(tensor, sp):
    head_cnt = tensor.size(2)
    remainder = head_cnt % sp
    if remainder != 0:
        pad_size = sp - remainder
        tensor_padded = torch.nn.functional.pad(
            tensor,
            pad=(0, 0, 0, pad_size, 0, 0, 0, 0),
            mode='constant',
            value=0.0
        )
        return tensor_padded
    else:
        return tensor

def unpad_heads(padded, ori_head_cnt):
    return padded[:, :, :ori_head_cnt, :]
\end{lstlisting}

\subsection{Communication Analysis}

We conducted communication complexity analysis on DeepSpeed-Ulysses \cite{jacobs2023deepspeed}, Ring-Attention \cite{liu2023ring}, USP \cite{fang2024usp}, Xtuner-Ulysses \cite{2023xtuner}, as well as our Dummy-Head-Ulysses approach. We uniformly set the batch size of the input data to $bs$, the original sequence length to $seq\_len$, and the hidden dimension to $d$, i.e., the initial input is $(bs, seq\_len, d)$. Our communication analysis accounts for the additional overhead introduced by sequence parallelism during both forward and backward propagation. The communication and time complexity comparison of all sequential parallel algorithms is shown in Table \ref{complexity_sp}.

\begin{table*}[h]
\centering
\caption{Complexity analysis of different sequence parallelism methods.}
\label{complexity_sp}
\begin{tabular}{lcc}
\toprule
\textbf{Method} & \textbf{Communication Complexity} & \textbf{Time Complexity} \\
\midrule
DeepSpeed-Ulysses & $O\left(\frac{8}{N} \times bs \times seq\_len \times d \right)$ & $O\left(bs \times seq\_len^2 \times \frac{d}{N}\right)$ \\
Ring-Attention & $O\left(4 \times bs \times seq\_len \times d\right)$ & $O\left(bs \times seq\_len^2 \times \frac{d}{N}\right)$ \\
USP & $O\left(\frac{8 + 4 sp_{cp}}{N} \times bs \times L \times d\right)$ & $O\left(bs \times seq\_len^2 \times \frac{d}{N}\right)$ \\
Xtuner-Ulysses & $O\left(\left(\frac{8}{N} + \frac{3}{insp}\right) \times bs \times seq\_len \times d\right)$ & $O\left(bs \times seq\_len^2 \times \frac{d}{N} \times insp\right)$ \\
Dummy-Head-Ulysses & $O\left(\frac{8}{N} \times \frac{hs_{new}}{hs} \times bs \times seq\_len \times d \right)$ & $O\left(bs \times seq\_len^2 \times \frac{d}{N} \times \frac{hs_{new}}{hs}\right)$ \\
\bottomrule
\end{tabular}
\end{table*}

\noindent{\textbf{DeepSpeed-Ulysses}.} The communication overhead in DeepSpeed-Ulysses originates from all-to-all operations, where query, key, value and output are exchanged during attention computation. This communication occurs twice, once during forward process and once during backward propagation. On modern clusters with intra-node NVSwitch interconnects inter-node fat tree IB topology, the total communication cost is: $O( \frac{8}{N} \times bs \times seq\_len \times d )$, where $N$ denotes the sequence parallel size. 

\noindent{\textbf{Ring-Attention.}} The communication in Ring-Attention arises from P2P communication of key and value during both forward and backward propagation in attention computation. The total communication cost is $O(4 \times bs \times seq\_len \times d)$. Since the sequence parallel size is usually greater than or equal to 2, Ring-Attention will have higher communication cost than DeepSpeed-Ulysses.

\noindent{\textbf{USP}.} We denote $sp_{cp}$ and $sp_{hp}$ as the sequence parallel size for Ring-Attention and DeepSpeed-Ulysses in USP, respectively. Then USP can be regarded as the outer layer performing Ring-Attention with sequence length $\frac{L}{sp_{hp}}$, and the inner layer performing DeepSpeed-Ulysses with sequence length $\frac{L}{sp_{cp}}$, so its final communication complexity is $O( \frac{8 + 4 sp_{cp}}{N} \times bs \times L \times d)$. Therefore, its communication complexity will introduce more time complexity compared to DeepSpeed-Ulysses. 

\noindent{\textbf{Xtuner-Ulysses.}} Xtuner-Ulysses' communication overhead originates from two components: \textit{all-gather} and \textit{all-to-all} operations. The \textit{all-to-all} communication complexity remains identical to DeepSpeed-Ulysses, while an additional \textit{all-gather} operation is required for query, key and value after \textit{all-to-all}, contributing $ O(\frac{3}{insp} \times bs \times seq\_len \times d)$. Thus, the total communication complexity is $O((\frac{8}{N} + \frac{3}{insp}) \times bs \times seq\_len \times d)$, where \textit{insp} represents the inner sequence parallel size. It is worth mentioning that, unlike the previous sequence parallelism, due to the operation of \textit{all\_gather}, the attention calculation complexity has doubled by the $insp$ times, which is caused by the increase of dim dimension.

\noindent{\textbf{Dummy-Head-Ulysses.}} The communication complexity of Dummy-Head-Ulysses is similar to that of Ulysses, except that we may add the head nums, so its communication complexity is $O( 8 \times \frac{hs_{new}}{hs} \times bs \times seq\_len \times \frac{d}{N})$. In most cases, we do not need to add head nums, and when we need to add it, in most models today, the head nums that needs to be added is usually very small, so the value of $\frac{hs_{new}}{hs}$ is only slightly greater than 1. In addition, due to the increase in head nums, the computational time complexity of the attention is $O\left(bs \times seq\_len^2 \times \frac{d}{N} \times \frac{hs_{new}}{hs}\right)$.

\section{Analysis of Sequence Parallelism}
In this section, we analyze several practical issues related to the application of sequence parallelism, including the details of distributed communication during training and the impact of position IDs.

\subsection{Distributed Communication Problem}
\begin{lstlisting}[caption={Sample code to verify distributed communication.}, label={dist_code}]
import torch
import torch.distributed as dist

USE_NN_REDUCE = 0

def main_worker(gpu):
    dist.init_process_group(
        backend="nccl", init_method="tcp://localhost:12345", world_size=2, rank=gpu
    )
    torch.cuda.set_device(gpu)
    w0 = torch.ones(1).cuda(device=gpu).requires_grad_()
    if dist.get_rank() == 0:
        x = torch.ones(1).cuda(device=gpu).requires_grad_() * 2
    else:
        x = torch.ones(1).cuda(device=gpu).requires_grad_() * 3
    
    y = torch.mul(w0, x)
    if USE_NN_REDUCE:
        y = dist.nn.all_reduce(y)
    else:
        dist.all_reduce(y)
    loss = 2 * y - 1
    loss.backward()

def local():
    w0 = torch.ones(1).cuda(device=0).requires_grad_()
    x = torch.ones(1).cuda(device=0).requires_grad_() * 5
    y = w0 * x
    loss = 2 * y - 1
    loss.backward()
\end{lstlisting}
Since all-reduce is used in sequence parallelism to aggregate the final loss, it is important to note that, in actual implementation, communication should be performed using $torch.distributed.nn.all\_reduce$ rather than $torch.distributed.all\_reduce$. This is because the latter does not implement the corresponding backpropagation wrapper, and the difference between the two can be found in \footnote{\url{https://github.com/pytorch/pytorch/issues/58005}}. We provide an additional code to analyze the difference between the two, as shown in Algorithm \ref{dist_code}.

Based on a simulated sequence parallel scenario using the test code above, we observe that the gradients of $w_0$ under $torch.distributed.nn.all\_reduce$ are 8 and 12 (GPU 0 and GPU 1), while those under $torch.distributed.all\_reduce$ are 4 and 6. These results differ by exactly the sequence parallel size, indicating incorrect scaling in the latter case. Furthermore, when sequence parallelism is not applied (i.e., in the local setting), the gradient of $w_0$ is expected to be 10, which matches the averaged result between two GPUs produced by $torch.distributed.nn.all\_reduce$. This consistency further validates the correctness of using $torch.distributed.nn.all\_reduce$ for gradient aggregation in sequence parallelism.

\begin{figure}[t]
  \centering
  \includegraphics[width=\columnwidth]{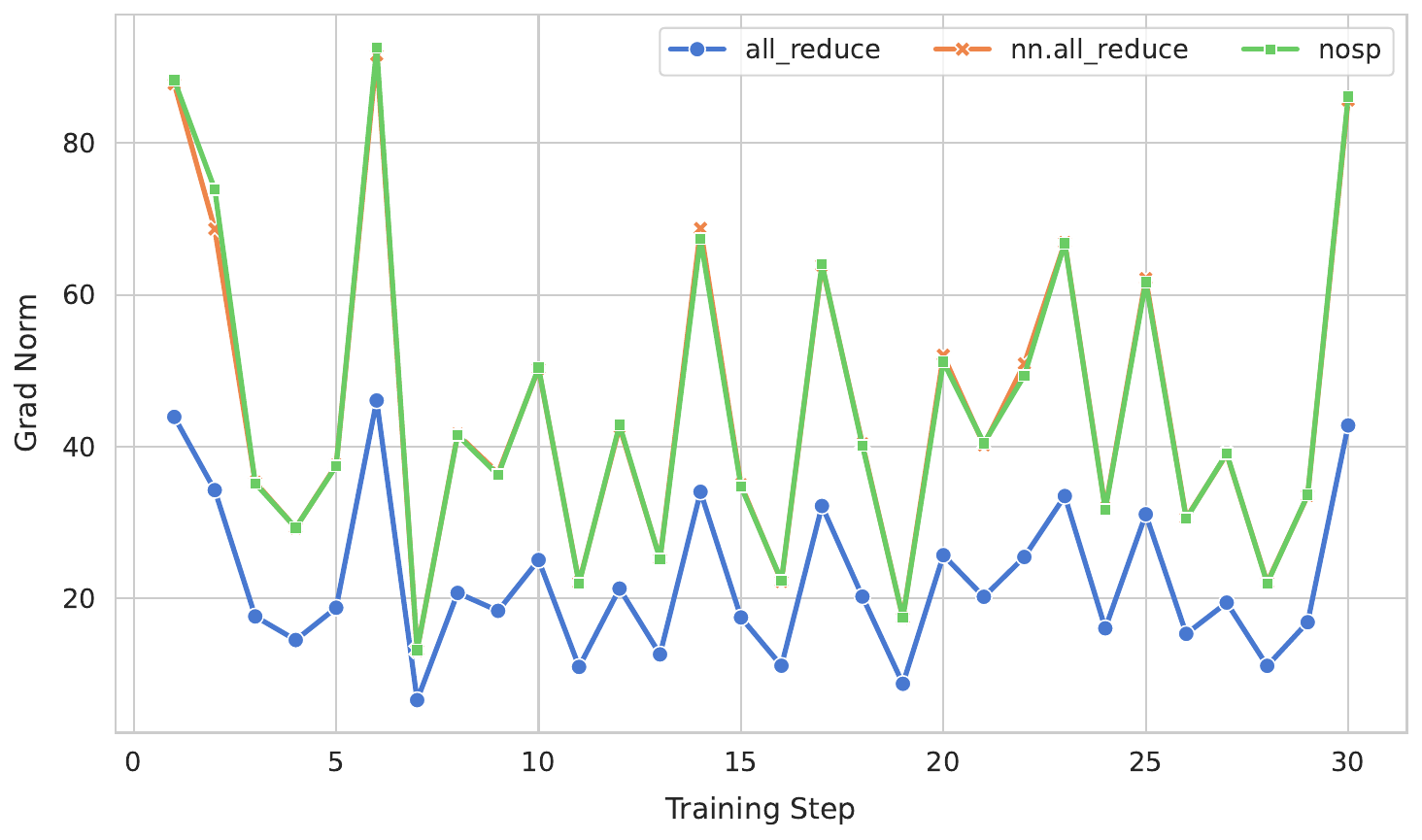}
  \caption{Performance of grad norm under distributed communication.}
  \label{fig:grad_norm_exp}
\end{figure}

We can perform the following analysis. Assume that the input data on two GPUs is $[x_0, x_1]$, and the intermediate computation yields $[y_0=w_0x_0, y_1 = w_0x_1]$. The final output on each GPU becomes $[y=y_0 + y_1, y=y_1 + y _0]$, leading to a loss of $[2y - 1, 2y - 1]$. Under $torch.distributed.all\_reduce$, the gradient with respect to $w_0$ becomes $[\frac{\partial loss}{\partial y}
\cdot \frac{\partial y}{\partial y_0}\cdot \frac{\partial y_0}{\partial w_0}, \frac{\partial loss}{\partial y}
\cdot \frac{\partial y}{\partial y_1}\cdot \frac{\partial y_1}{\partial w_0}]$. In contrast, $torch.distributed.nn.all\_reduce$ also performs an $all reduce$ operation during the backward pass, which results in a final gradient proportional to $[all
\_reduce(\frac{\partial loss}{\partial y} )
\cdot \frac{\partial y}{\partial y_0}\cdot \frac{\partial y_0}{\partial w_0}, all
\_reduce(\frac{\partial loss}{\partial y} )
\cdot \frac{\partial y}{\partial y_1}\cdot \frac{\partial y_1}{\partial w_0}]$, which differs from the former by a factor equal to the sequence parallel size. We also evaluated the impact of this issue within our framework. As shown in Figure \ref{fig:grad_norm_exp}, the gradient norm differs by a factor corresponding to the sequence parallel size. 

\subsection{The Impact of Position IDs}
It should be noted that most current decoder-only models, such as Qwen \cite{yang2024qwen2}, LLaMa \cite{grattafiori2024llama}, etc., use RoPE \cite{su2024roformer} position encoding. When using sequence parallelism, if the $position\_ids$ parameter is not explicitly passed in, the existing sequence will be re-encoded on each GPU, which will cause serious errors. That is, when no $position\_ids$ is passed in, for a sequence of length $seq\_len$, it is split into $seq\_len / sp$ length on each GPU, and it will be assigned the default $position\_ids$ of $[0, 1, 2, ..., seq\_len / sp - 1]$, Where $seq\_len$ is the length of the training data and $sp$ is the sequence parallel size. However, since it essentially represents a complete sequence, it should have a complete position encoding, that is, its position encoding should be a partition of $[0, 1, 2, ... seq\_len - 1]$. Therefore, we need to initialize the $position\_ids$ of the data in advance and explicitly pass them into the forward process of the corresponding model.



\section{Experiments}
The experimental section focuses primarily on verifying the correctness of our implementation, as well as comparing the performance of DeepSpeed-Ulysses and Ring-Attention, including their maximum supported sequence lengths and throughput efficiency. In addition, we provide an experimental analysis of the performance and efficiency of our proposed Dummy-Head-Ulysses variant.

\subsection{Correctness Verification}

\textbf{Experiment Settings.} To verify correctness, we construct 30 samples each for the SFT and DPO tasks. The experiments were conducted using the Qwen2.5-0.5B-Instruct \cite{yang2024qwen2} model. For training, we used a gradient accumulation step of 8, trained for 8 epochs, and set the input sequence length to 8k tokens. We adopted DeepSpeed ZeRO Stage 3 with offloading, enabled sequence parallel size 2, and used bfloat16 precision. The learning rate was set to $5 \times 10^{-5}$ for SFT and $1\times 10^{-6}$ for DPO, with the DPO $\beta$ parameter set to 0.1. When sequence parallelism was enabled, we used 2×A100 GPUs (80GB); otherwise, we used a single A100 (80GB).

\begin{figure}[t]
  \centering
  \includegraphics[width=\columnwidth]{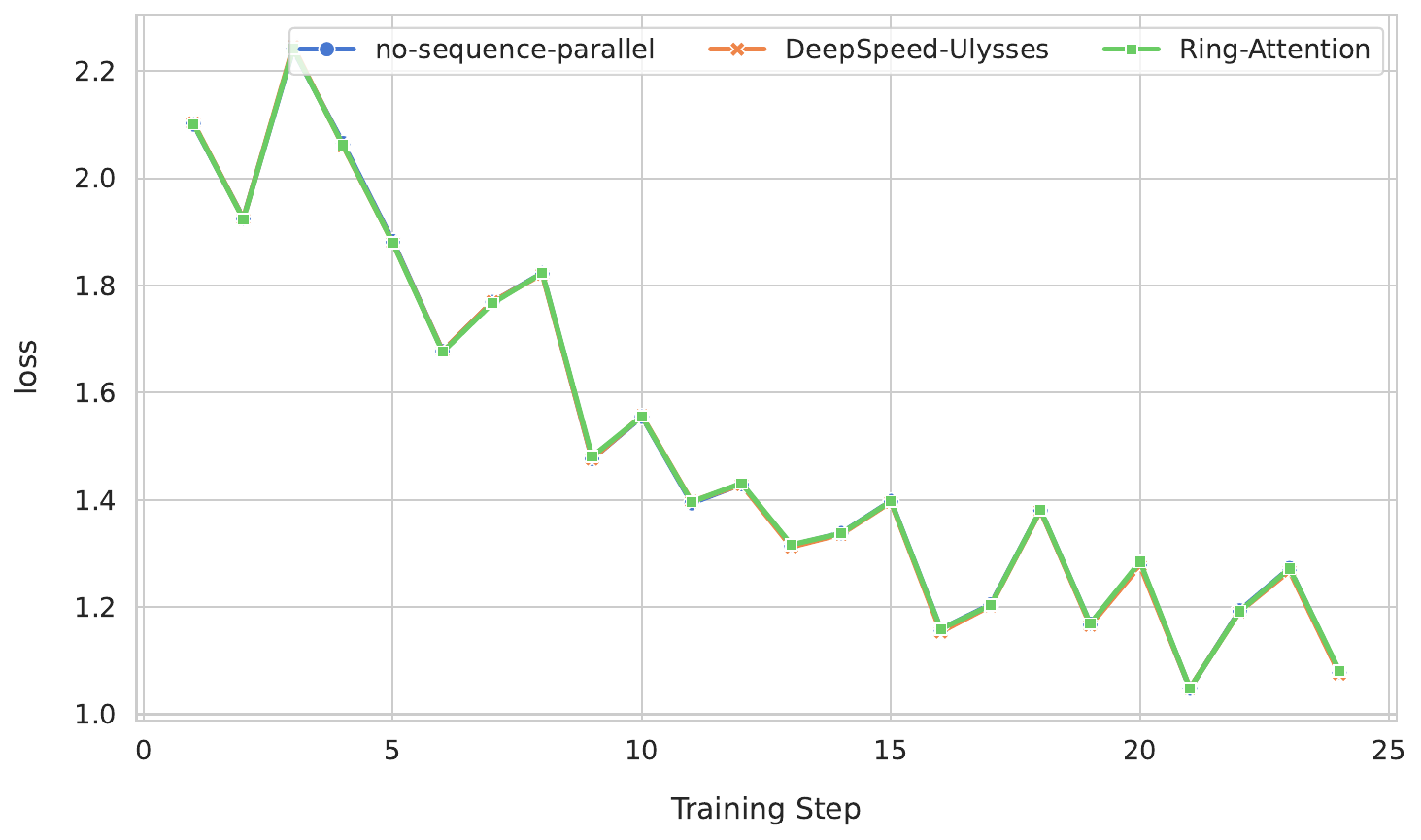}
  \caption{Comparison of SFT loss with and without SP.}
  \label{fig:sft_correct}
\end{figure}

\begin{figure}[t]
  \centering
  \includegraphics[width=\columnwidth]{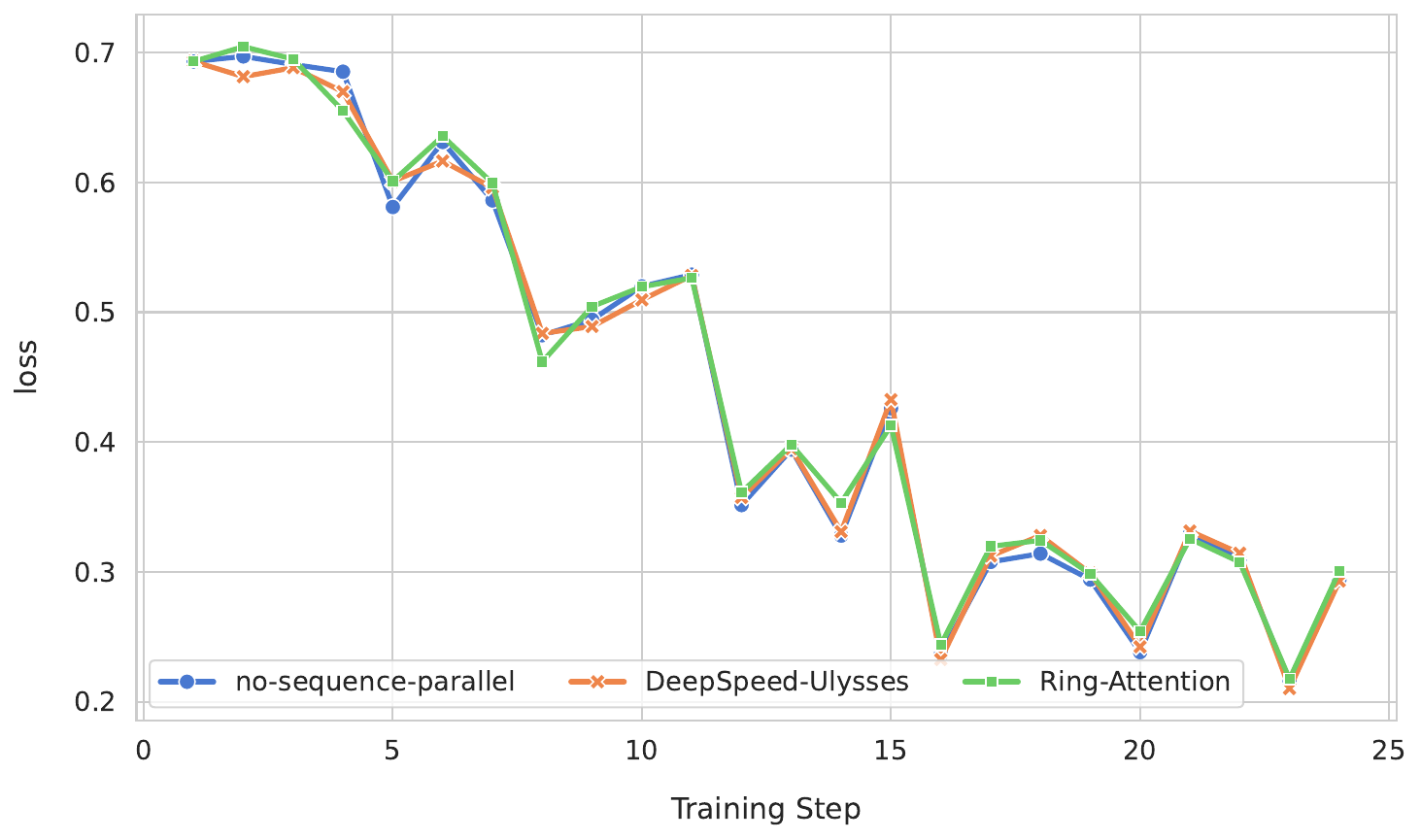}
  \caption{Comparison of DPO loss with and without SP.}
  \label{fig:dpo_correct}
\end{figure}

\textbf{Results.} The experimental results are presented in Figures \ref{fig:sft_correct} and \ref{fig:dpo_correct}. As shown, both DeepSpeed-Ulysses and Ring-Attention under our implementation produce loss curves that are nearly identical to those obtained without sequence parallelism.

For SFT, the loss difference is negligible, with the curves almost perfectly overlapping. In the case of DPO, the loss exhibits slightly greater variance. This is partially due to the inherently smaller magnitude of DPO loss values. Furthermore, we conducted a experiment in which the learning rate was set to zero, effectively disabling parameter updates, to isolate the effect of backward communication. In this setting, the $\pi_\theta(y_w|x)$ and $\pi_\theta(y_l|x)$ from the forward pass matched exactly between the sequence-parallel and non-sequence-parallel implementations, indicating that any discrepancies originate from the backward communication in sequence parallelism.

\begin{table}[htbp]
    \centering
    \caption{Max sequence length comparison between DeepSpeed-Ulysses (DU) and Ring-Attention (RA).}
    \label{tab:max_length}
    \begin{tabular}{llcrr}
        \toprule
        Model & Method & SP & DU & RA \\
        \midrule
        \multirow{4}{*}{Qwen2.5-7B} 
            & \multirow{2}{*}{SFT} & 4 & 86k & \textbf{96k} \\
            &                      & 8 & 166k & \textbf{182k} \\
            & \multirow{2}{*}{DPO} & 4 & \textbf{38k} & 34k \\
            &                      & 8 & \textbf{76k} & 60k \\
        \midrule
        \multirow{2}{*}{Qwen2.5-14B}
            & SFT & 8 & 136k & \textbf{152k} \\
            & DPO & 8 & \textbf{68k} & 52k \\
        \midrule
        \multirow{2}{*}{Qwen2.5-72B}
            & SFT & 8 & \textbf{132k} & 110k \\
            & DPO & 8 & \textbf{46k} & 44k \\
        \bottomrule
    \end{tabular}
\end{table}

\begin{table*}[htbp]
\centering
\caption{Throughput (Tokens/s) Comparsion between differnt sequence parallel methods.} 
\label{tab:performance}
\begin{tabular}{lllrrrrrr}
\toprule
Model & Method & Length & Ulysses & DHU & XU & USP-u4 & USP-u2 & RA  \\
\midrule
1.5B  & SFT & 128k & -- & 1351.66
 & 1131.08 & 1720.42 & \textbf{1809.38} & 1732.45  \\
1.5B  & DPO & 32k & -- & 1256.85 & 1070.64 & \textbf{1758.42} & 1637.75 & 1421.16  \\
3B  & SFT & 100k & \textbf{1118.50} & -- & -- & 1027.85 & 919.56 & 890.52  \\
3B  & DPO & 32k & 753.13 & -- & -- & \textbf{951.60} & 885.27 & 885.27 \\
7B  & SFT & 130k & -- & \textbf{619.88} & 380.86 & 594.33 & 586.55 & 567.27  \\
7B  & DPO & 48k & -- & 570.62 & 426.70 & \textbf{709.51} & 615.36 & 689.31  \\
14B & SFT & 100k & \textbf{284.73} & -- & -- & 281.94 & 282.51 & 282.78 \\
14B & DPO & 32k & 250.48 & -- & -- & \textbf{321.44} & 272.75 & 307.44  \\
32B & SFT & 80k & 146.55 & -- & -- & \textbf{156.15} & 141.57 & 139.55  \\
32B & DPO & 24k & 104.10 & -- & -- & \textbf{141.80} & 120.67 & 126.90  \\
\bottomrule
\end{tabular}
\end{table*}




\subsection{Performance camparsion of DeepSpeed-Ulysses and Ring-Attention}
This subsection compares the maximum supported sequence length and runtime efficiency of DeepSpeed-Ulysses and Ring-Attention.

\textbf{Experiment Settings.} We constructed SFT and DPO datasets in which each individual sample exceeds 200k tokens. This design ensures that the evaluation of maximum sequence length is not affected by padding, thereby providing accurate stress-testing results\footnote{All experiments were conducted using torch 2.2.1, transformers 4.45.2 \cite{wolf2020transformers} and flash\_attention 2.6.1.}. The batch size was fixed to 1 across all experiments, with sequence parallel size set to either 4 or 8. We employed DeepSpeed ZeRO Stage 3 with offloading and trained using bfloat16 precision.
Length and runtime efficiency evaluations were conducted on Qwen2.5-7B, Qwen2.5-14B, and Qwen2.5-72B. The 72B model was trained on 32×A100 (80GB) GPUs, while the other models were trained on 8×A100 (80GB) GPUs. The learning rate was set to $5 \times 10^{-6}$ for SFT and $1 \times 10^{-6}$ for DPO, with the DPO beta parameter set to 0.1.

\textbf{Results.} The results of maximum sequence length stress testing are presented in Table \ref{tab:max_length}. It can be observed that sequence parallelism enables both SFT and DPO to process longer sequences under limited resource conditions. However, due to the additional communication overhead introduced by sequence parallelism, the improvement in maximum sequence length is not strictly proportional to the increase in the number of devices.

Furthermore, when comparing the two implementation approaches, we find that DeepSpeed-Ulysses generally supports longer sequences in DPO tasks, while it tends to support shorter lengths in SFT. This difference may stems from DPO's additional reference model, which introduces extra communication overhead such that DeepSpeed-Ulysses appears more efficient. 
The suboptimal performance of Ring-Attention on the 72B model may be attributed to the additional communication overhead across multiple nodes, which leads to a reduction in the maximum sequence length achievable for both SFT and DPO tasks. These observations suggest that the choice of sequence parallelism strategy should be guided by the characteristics of the specific training task.

\subsection{Throughput Comparsion of Different Sequence Parallelism Methods}



To evaluate the effectiveness of our Dummy-Head Ulysses (DHU) implementation, we compare its throughput with Xtuner-Ulysses (XU), Ulysses, USP-u(s, the sequence number of ulysses degree) and Ring-Attention (RA).

\textbf{Experiment Settings.} We conduct experiments using different size qwen2.5-models with a sequence parallel size of 8, where only the 1.5B and 7B models encounter cases where the number of attention heads is not divisible by the sequence parallel size. The dataset and hyperparameters are consistent with those used in Section 6.2. All experiments are conducted on 8 × A100 (80GB) GPUs. 

\textbf{Results.} The experimental results for both SFT and DPO tasks are shown in Table \ref{tab:performance}.
It can be observed that our Dummy-Head-Ulysses achieves higher throughput compared to the Xtuner-Ulysses. This improvement is attributed to the fact that Xtuner-Ulysses replicates the same attention heads across multiple devices, which leads to increased memory consumption and computational overhead. In contrast, our method only introduces a small number of additional heads as padding.

Our Dummy-Head-Ulysses exhibits marginally lower throughput than USP in certain cases (e.g., DPO with 7B models), likely attributable to increased attention computation complexity introduced by the dummy-head mechanism. While DeepSpeed-Ulysses maintains optimal throughput in most scenarios, we observe that USP-U4 achieves superior throughput in selected experiments. Notably, while DeepSpeed-Ulysses generally achieves optimal throughput, USP-U4 outperforms it in some cases—with DeepSpeed-Ulysses even underperforming Ring-Attention. This stems from its reliance on \textit{flash\_attn\_varlen\_func} (triggered by passing in \textit{attention\_mask} for neat-packing compatibility), which incurs overhead in padding-free scenarios like our experiments, though it benefits padded data.

\section{Conclusion and Future Work}

In this work, we present the integration of sequence parallelism strategies into the LLaMA-Factory framework to support long-sequence training. We provide a detailed account of the implementation process and key challenges. In addition, we proposed Dummy-Head-Ulysses to solve the problem that the head nums cannot divide the sequence parallel size encountered by DeepSpeed-Ulysses. We verified the accuracy of our implementation through experiments and carefully analyzed a series of indicators such as the max sequence length and throughput of different sequence parallel methods.



In future work, we plan to continuously improve our repository with a focus on the following directions:

(1) Extending support to a broader range of models, including enabling sequence parallelism for multimodal models;

(2) Exploring more efficient sequence parallelism strategies to further reduce memory consumption and improve computational efficiency;

(3) Enhancing functionality to support more efficient training workflows, such as enabling precomputation of reference model outputs in DPO.

\section*{Limitations}

The limitations of our current work are summarized as follows:

(1) In extending DeepSpeed-Ulysses, the implementation of Dummy-Head-Ulysses incurs a certain amount of overhead. We aim to explore more efficient implementations to ease this cost.

(2) We observe that the DPO loss exhibits a small discrepancy. While the deviation remains within an acceptable range, we are interested in investigating approaches to further reduce this error.



\bibliography{custom}




\end{document}